\documentclass[sigconf]{acmart} 

\usepackage{amsmath,amsfonts,bm}

\def\eqref#1{equation~\ref{#1}}

\def\1{\bm{1}}

\DeclareMathAlphabet{\mathsfit}{\encodingdefault}{\sfdefault}{m}{sl}
\SetMathAlphabet{\mathsfit}{bold}{\encodingdefault}{\sfdefault}{bx}{n}

\usepackage{hyperref}
\usepackage{url}
\usepackage{graphicx}
\usepackage{booktabs}
\usepackage{multirow}
\usepackage{subcaption}
\usepackage{enumitem}
\usepackage{tabularx}

\AtBeginDocument{%
  \providecommand\BibTeX{{%
    \normalfont B\kern-0.5em{\scshape i\kern-0.25em b}\kern-0.8em\TeX}}} 

\setcopyright{acmlicensed}
\copyrightyear{2018}
\acmYear{2018}
\acmDOI{XXXXXXX.XXXXXXX}

\acmConference[KDD '24]{the 30th ACM SIGKDD
Conference on Knowledge Discovery and Data Mining}{August25-–29}{Barcelona, Spain}
 
\acmISBN{978-1-4503-XXXX-X/18/06}

\begin{document}

\newcommand{\VZ}[1]{\textcolor{red}{[Vincent: #1]}}
\newcommand{\dc}[1]{\textcolor{orange}{[Didier: #1]}}
\newcommand{\wenyi}[1]{\textcolor{green}{[wenyi: #1]}}
\newcommand{\yx}[1]{\textcolor{magenta}{[yingxue: #1]}}
\newcommand{\huiling}[1]{\textcolor{blue}{[Huiling: #1]}}

\title{GraSS: Combining Graph Neural Networks with Expert Knowledge for SAT Solver Selection}

\author{Zhanguang Zhang}
\affiliation{%
    \institution{Huawei Noah's Ark Lab}
    \city{Montreal}
    \country{Canada}
}
\email{zhanguang.zhang@huawei.com}

\author{Didier Chetelat}
\affiliation{
    \institution{Huawei Noah's Ark Lab}
    \city{Montreal} \country{Canada}
}

\author{Joseph Cotnareanu}
\affiliation{
    \institution{McGill University}
    \city{Montreal} \country{Canada}
}

\author{Amur Ghose}
\affiliation{
    \institution{Huawei Noah's Ark Lab}
    \city{Montreal} \country{Canada}
}

\author{Wenyi Xiao}
\affiliation{
    \institution{Huawei Noah's Ark Lab}
    \city{Hong Kong} \country{China}
}

\author{Hui-Ling Zhen}
\affiliation{
    \institution{Huawei Noah's Ark Lab}
    \city{Hong Kong} \country{China}
}

\author{Yingxue Zhang}
\affiliation{
    \institution{Huawei Noah's Ark Lab}
    \city{Montreal} \country{Canada}
}

\author{Jianye Hao}
\affiliation{
    \institution{Huawei Noah's Ark Lab}
    \city{Beijing} \country{China}
}

\author{Mark Coates}
\affiliation{
    \institution{McGill University}
    \city{Montreal} \country{Canada}
}

\author{Mingxuan Yuan}
\affiliation{
    \institution{Huawei Noah's Ark Lab}
    \city{Hong Kong} \country{China}
}

\renewcommand{\shortauthors}{Zhang, et al.}

\begin{abstract}
Boolean satisfiability (SAT) problems are routinely solved by SAT solvers in real-life applications, yet solving time can vary drastically between solvers for the same instance. This has motivated research into machine learning models that can predict, for a given SAT instance, which solver to select among several options. Existing SAT solver selection methods all rely on some hand-picked instance features, which are costly to compute and ignore the structural information in SAT graphs. In this paper we present GraSS, a novel approach for automatic SAT solver selection based on tripartite graph representations of instances and a heterogeneous graph neural network (GNN) model. While GNNs have been previously adopted in other SAT-related tasks, they do not incorporate any domain-specific knowledge and ignore the runtime variation introduced by different clause orders. We enrich the graph representation with domain-specific decisions, such as novel node feature design, positional encodings for clauses in the graph, a GNN architecture tailored to our tripartite graphs and a runtime-sensitive loss function. Through extensive experiments, we demonstrate that this combination of raw representations and domain-specific choices leads to improvements in runtime for a pool of seven state-of-the-art solvers on both an industrial circuit design benchmark, and on instances from the 20-year Anniversary Track of the 2022 SAT Competition.
\end{abstract}

\ccsdesc[500]{Computing methodologies~Neural networks}
\ccsdesc[500]{Computing methodologies~Cost-sensitive learning}
\ccsdesc[500]{Mathematics of computing~Graph algorithms}

\keywords{GNN, SAT, Algorithm Selection}

\maketitle

\section{Introduction}

\begin{figure*}[t]
    \centering
    \includegraphics[width=\linewidth, trim={0.2cm, 0, 0, 0}, clip]{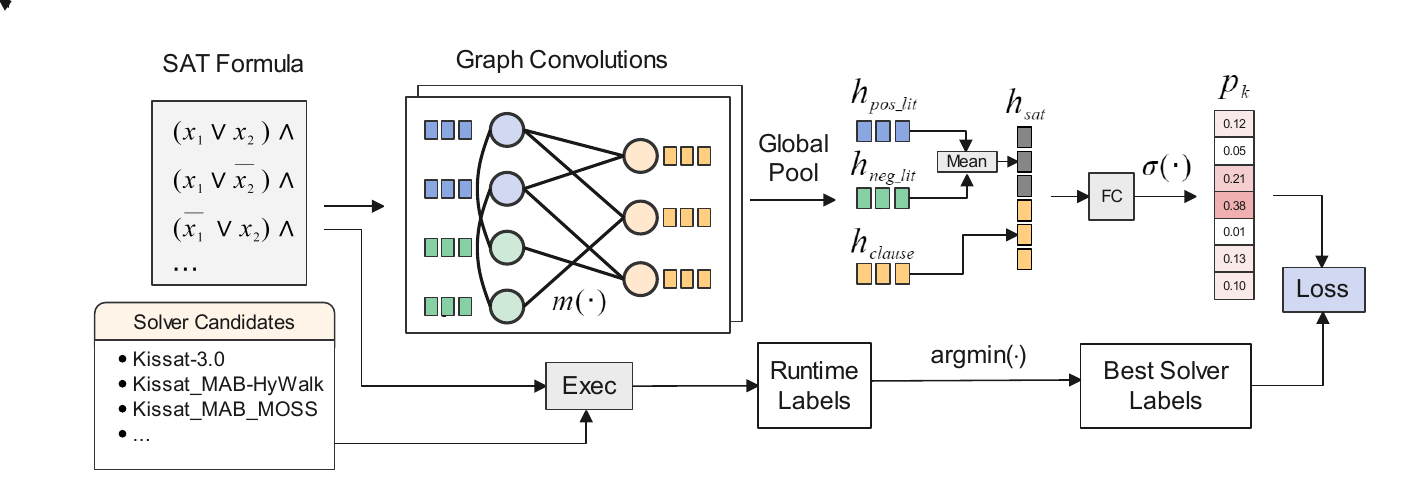}
    \caption{The workflow of our method. SAT instances are represented as literal-clause graphs with hand-designed attributes. Rounds of heterogeneous graph convolutions are applied, which modify the attributes. The attributes of the clause and variable nodes are then averaged, before being fed to a linear layer followed by a softmax over the various solvers. The convolutions and the linear layer are trained to minimize by gradient descent a runtime-sensitive classification loss computed from runtimes collected on training SAT instances.} 
    \label{fig:workflow}
\end{figure*}

The Boolean satisfiability (SAT) problem  is one of the most fundamental computer science problems, with numerous applications in planning and scheduling \citep{kautz1996encoding, crawford1994experimental}, formal software verification \citep{ganai2007sat, ivanvcic2008efficient} and electronic circuit design \citep{goldberg2001using, huang2012formal}. A SAT instance consists of a formula with Boolean variables, such as $(x_1 \vee \overline{x}_2) \wedge (\overline{x}_1 \vee x_2 \vee x_3) \wedge \overline{x}_1$, and the problem involves finding an assignment of values for each variable $x_i$ which makes the whole formula true, or proving that no such assignment exists. Although the problem is NP-complete \citep{cook2023complexity}, many SAT solvers have been designed over the years and modern CDCL-based solvers are routinely able to solve industrial problems within minutes \cite{marques2021conflict}.

Structural differences between different SAT problems mean that the choice of solver can have a dramatic impact on the solving time. This has motivated the use of machine-learning based methods for selecting the optimal solver to use for a given instance, with the hope that data-driven models can see patterns where humans have been unsuccessful. The most influential of those has probably been SATzilla  \citep{xu2008SATzilla}, which has won several times the annual SAT Competition \citep{satcompetition}. This model relies on machine learning algorithms that require a fixed-dimensional vector of features as input, irrespective of the actual instance size (number of clauses and variables). This necessarily implies that some aspects of a SAT problem are not taken into account when performing solver selection.

In recent years, machine learning has been revolutionized by deep learning models trained on raw descriptions of data points such as image pixels or text strings \citep{GoodBengCour16}. In particular, surprising success has been found in a variety of combinatorial optimization tasks by representing optimization problems as graphs, and feeding them as inputs to graph neural networks \citep{cappart2023combinatorial}. Such models are able to take the complete representation of a problems as input, in a size-independent way, and see patterns where humans have been unable to distinguish any.

In this work, we propose \textbf{GraSS} (\textbf{Gra}ph Neural Network \textbf{S}AT Solver \textbf{S}elector), the first graph neural network (GNN) based method for automatic SAT solver selection. We represent instances as literal-clause graphs \citep{guo2023machine}, thus encoding the entirety of the information pertaining to an instance. To improve performance further, we also endow the graph with hand-designed features representing domain-knowledge about which aspects of the graph should be particularly useful for solver selection, as well as positional encodings for the clauses to allow for order-specific effects. Our GNN model consists of learned graph convolutions operating over each type of edge, with a node-specific pooling operation prior to a linear classifier. We train our model in a supervised manner with a runtime-sensitive classification loss. The training data consists of a collection of instances for which the runtimes of multiple solvers have been collected.

On both a large-scale industrial circuit design benchmark, and on instances from the Anniversary Track of the 2022 SAT Competition \citep{satcompetition2022}, we report improvements in performance compared to seven competitive solvers, as well as state-of-the-art machine learning approaches. We also perform a complete ablation study to rigorously test the importance of each component of our proposed pipeline.

In summary, our contributions are as follows.
\begin{itemize}[leftmargin=*]
\item We propose the first approach for SAT solver selection that makes complete use of the SAT instance data, by representing each instance as an attributed graph and using a GNN model.
\item We propose a model architecture that is tailored to the tripartite graph representations.
\item We design novel node-level features to incorporate domain-specific knowledge. 
\item We report for the first time the value of including positional encodings for clauses in the graphical representation of an instance for a SAT-related machine learning task.
\item We introduce a novel runtime-sensitive classification loss, which could be of value for general algorithm selection tasks.
\item We report state-of-the-art empirical performance on two hard SAT benchmarks and conduct extensive ablation studies to confirm the value of our architectural choices.
\end{itemize}
Collectively, these elements strongly suggest our approach should be regarded as a new  standard in the field of SAT solver selection.
\section{Related Work}\label{sec:literature}

\subsection{SAT Solver Selection}

There exists a rich literature describing machine learning models for the selection of the optimal SAT solver for a given instance. This approach is sometimes referred to as portfolio-based SAT solving. A detailed summary is provided by \citet[Section 6]{holden2021machine}.

SATzilla \citep{xu2008SATzilla} and its successors \citep{xu2009satzilla, xu2012satzilla} are a family of classification models that have won multiple prizes in the SAT Competition (2007 and 2009) and the SAT Challenge (2012). SATzilla uses hand-selected features to characterize each SAT instance for best solver selection. The latest version of SATzilla \citep{xu2012satzilla} consists of a feature cost classifier to predict if the entire set of features can be computed within a time threshold and an algorithm selector for instances with feature cost less or equal to this threshold. Whereas the first version \citep{xu2008SATzilla} applies ridge regression, the latest version \citep{xu2012satzilla} uses a cost-sensitive decision tree for the algorithm selection model.

In addition to the SATzilla family, hand-picked features have also been used with $k$-nearest neighbor or clustering methods to select the best algorithm \citep{nikolic2013argosmart, kadioglu2011algorithm, malitsky2013boosting}. CSHC \citep{malitsky2013boosting} uses a cost-sensitive hierarchical clustering algorithm to iteratively partition the feature space into clusters in a supervised top-down fashion. In combination with the static algorithm schedule of 3S \cite{kadioglu2011algorithm}, it achieved better performance than SATzilla in the 2011 SAT Competition. ArgoSmArT \cite{nikolic2013argosmart} uses a combination of a $k$-Nearest Neighbors (KNN) model and a multi-armed bandit algorithm. The KNN performs solver selection based on the nearest instance to the queried instance in a metric space.

Finally, \citet{loreggia2016deep} proposed a deep learning approach based on convolutional neural networks (CNNs). They take textual representations of SAT instances in a standard format, and convert them into grayscale images by replacing each character with its corresponding ASCII code. These images are then rescaled to 128x128 pixels, and fed to a CNN, which is trained to predict the fastest solver in the pool. Although employing deep learning, this approach is not lossless, as the rescaling required by the CNN architecture leads to information loss, unlike our approach.

\subsection{Algorithm Selection}

More generally, SAT solver selection is a special case of algorithm selection, which aims to select, for a given input, the most efficient algorithm from a set of candidate algorithms. This is particularly important for computationally hard problems, where there is typically no single algorithm that outperforms all others for all inputs. Besides SAT solving, algorithm selection techniques have achieved remarkable success in various applications such as Answer Set Programming (ASP) \citep{gagliolo2011gambleta} and the Traveling Salesperson Problem (TSP) \citep{applegate2007tsp}. A thorough literature review is provided in \citet{kerschke2019automated}.

Algorithm selection methods can be roughly divided between offline and online methods. Offline methods, such as this work, rely on training ahead of time on a labeled dataset, whereas online algorithms attempt to improve selection performance as more and more cases are run. Although providing worse initial performance, online methods avoid the computational cost of the initial training phase and the problem of distribution shift between training and test data, and methods based on based on reinforcement learning or multi-armed bandits have been proposed for this purpose \citep{degroote2016reinforcement, gagliolo2011gambleta}. 

The many design choices in algorithm selection systems pose new challenges for efficient system development. For example, AutoFolio \citep{lindauer2015autofolio} automatically configures the entire framework, including budget allocation for pre-solving schedules, pre-processing procedures (such as transformations and filtering) and algorithm component selection. It achieved competitive performance in multiple scenarios from the ASLib \citep{bischl2016aslib} benchmark.

\subsection{Graph Neural Networks in SAT Solving} \label{tag:sat_repr}

Several works have explored applications of GNNs to various aspects of SAT solving in the past, even if not specifically to the problem of solver selection. In all these works, some kind of graphical representation of SAT instances is used. Multiple suggestions have appeared in the literature: these include lossless representations like literal-clause graphs (LCGs) and variable-clause graphs (VCGs), and lossy representations like literal-incidence graphs (LIGs) and variable-incidence graphs (VIGs) \citep{guo2023machine}. These different representations strike a balance between graph size and information content, and have found success in various SAT-related tasks.

Some prior works have explored the use of GNNs to learn local search heuristics in SAT solvers \cite{yolcu2019nips,kurin2020nips}. \citet{yolcu2019nips} represent SAT formulas as variable-clause graphs (VCGs) and a GNN model is trained to select variables whose sign to flip at every step through a Markov decision process (MDP). The learned heuristic is shown to reduce the number of steps required to solve the problem. Graph-Q-SAT \citep{kurin2020nips} uses a deep Q-network (DQN) with GNN architecture to learn branching heuristics in conflict driven clause learning (CDCL) solvers. Each SAT formula is converted into a VCG, and GNN layers are used to predict the $Q$-value of each variable. The variable with the highest $Q$-value for the specific assignment is selected for branching. The learned heuristic is shown to significantly reduce the number of iterations required for SAT solving.

Instead of relying on existing SAT solvers, NeuroSAT \citep{selsam2019neurosat} uses a GNN-based model to predict satisfiability of an instance in an end-to-end manner. Each SAT formula is represented by a literal-clause graph (LCG), as in our work. 
After several steps of message-passing, the updated embedding of each literal is projected to a scalar ``vote'' to indicate the confidence that the formula is satisfiable. The votes are averaged together and passed through a sigmoid function to produce the model's probability that the instance is satisfiable. On randomly generated instances from a SR(40) distribution, NeuroSAT solved 70\% of SAT problems with an accuracy of 85\%. A subsequent work, NeuroCore \citep{selsam2019guiding}, uses a lighter NeuroSAT model to predict the ``core'' of instance, which is the smallest unsatisfiable subset of clauses. This prediction is then used to guide variable selection in SAT solver algorithms.

Finally, graph neural networks have also been widely used for SAT instance generation. For example, G2SAT \citep{you2019g2sat} and HardSATGEN \citep{li2023hardsatgen} represent instances as LCGs, and generate new variants from an iterative splitting and merging process driven by a GNN. Furthermore, W2SAT \citep{wen2023w2sat} extends this approach by representing instances as weighted graphs encoding literal co-occurence among clauses, while using a similar generation mechanism.

\section{Approach}\label{sec:methodology}

We now describe our proposed approach. The workflow of our method is shown in Figure~\ref{fig:workflow}.

\subsection{Problem}
Our SAT solver selection problem can be formally described as follows. We have a fixed pool of SAT solvers $S_1, \dots, S_K$ and a SAT instance $a$, and we must select a solver so as to solve the instance in the smallest runtime possible.

To do so, at train-time, we are given a collection of SAT instances and for each instance $a_i$, we are given sampled solving times $t_i^1, ..., t_i^K$ from each solver.  This dataset can be used for training a machine learning model offline in a supervised manner. At test-time, a separate collection of SAT instances are given to the model, which selects a solver for each. The selected solver for each instance is run, and the average runtime over all testing instances is used to benchmark performance of the selector model.

\subsection{Representation and features}\label{sec:representation}

The inputs to the model are individual SAT instances. We assume that they are formulated in conjunctive-normal form (CNF), that is as a formula $c_1\wedge \dots\wedge c_m$ where each $c_i$ is a \emph{clause}, which in turn is in the form $c_i=l_1\vee l_2\vee \dots$, where each \emph{literal} $l_i$ stands for a variable $x_j$ in the problem, or its negation $\bar{x}_j$. Any SAT problem can be converted into an equivalent CNF problem in linear time \citep{tseitin1983complexity}.

\begin{figure}
    \includegraphics[width=\linewidth, trim={0.05\linewidth, 0, 0, 0}, clip]{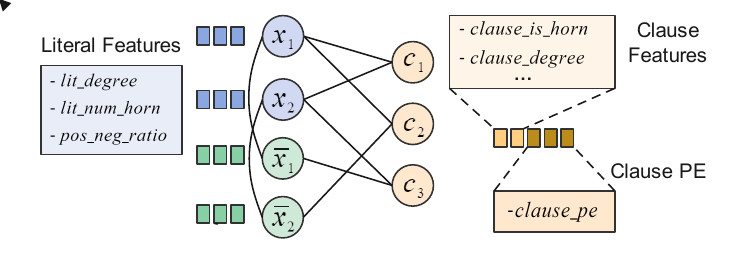}
    \caption{Literal-clause graph representation of a SAT instance used in this work. The instance is converted to CNF form, and nodes represent positive literals, negative literals and clauses. Edges are drawn between clauses and literal nodes if the literal participate in the clause, and edges are also drawn between positive and negative nodes of the same variable. The nodes are endowed with feature vectors described in Appendix~\ref{sec:node-features}.}
    \label{fig:lcg}
\end{figure}

Given CNF SAT instances, we input them to the machine learning model as literal-clause graphs (LCGs) \citep{guo2023machine} endowed with extra information in the form of node features. A representation is provided in Figure \ref{fig:lcg}. The literal-clause graph of a SAT instance is an undirected graph with three types of node: one node $c_j$ per clause in the graph, and two nodes $x_i$ and $\bar{x}_i$ per variable in the graph, representing itself and its negation respectively. An edge is drawn between a clause node $c_j$ and a variable node $x_i$/$\bar{x}_i$ if the variable (respectively, its negation) appears as literal in the clause. Finally, an edge is drawn between every positive and negative variable node.

In addition, to every clause node and variable node, we attach feature vectors. Most features are hand-designed and are inspired by those used by SATzilla \cite{xu2009satzilla, xu2012satzilla}. They represent expert knowledge that is known to be critical for SAT solving process, such as the presence of Horn clauses, which are clauses containing at most one positive literal. These are especially important for the solving process as the collection of Horn clauses can be proved within linear time \citep{dowling1984linear}. Besides these hand-designed features, clause node features are also enriched with a positional encoding, described in the next subsection. A complete list of the features used is provided in Appendix~\ref{sec:node-features}.

\subsubsection{Clause positional embeddings}\label{sec:clause_pe}

\begin{figure*}
    \centering
    \begin{subfigure}{0.5\linewidth}
        \centering
        \includegraphics[width=0.95\linewidth]{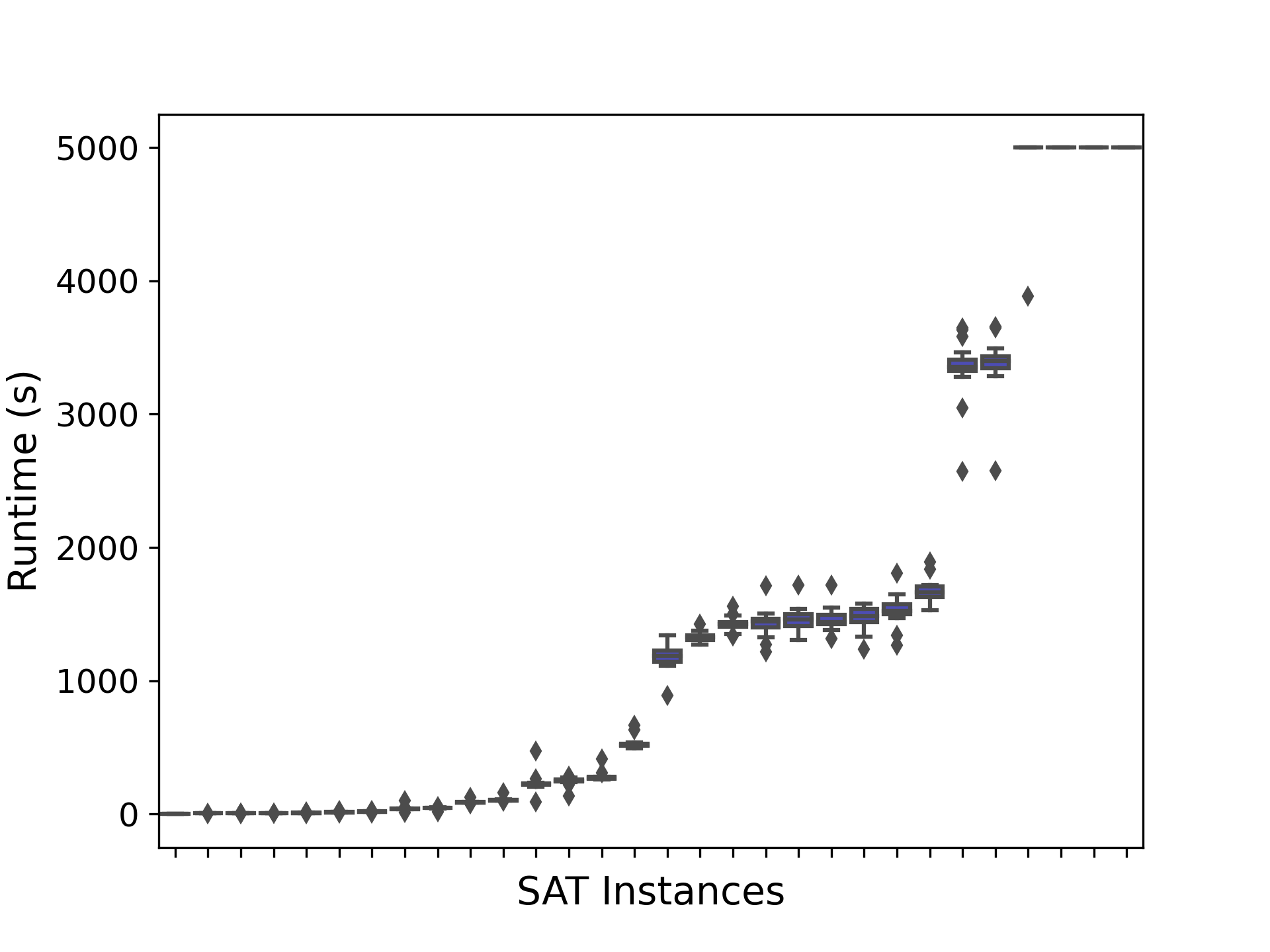}
    \end{subfigure}%
    \hfill
    \begin{subfigure}{0.5\linewidth}
        \centering
        \includegraphics[width=0.95\linewidth]{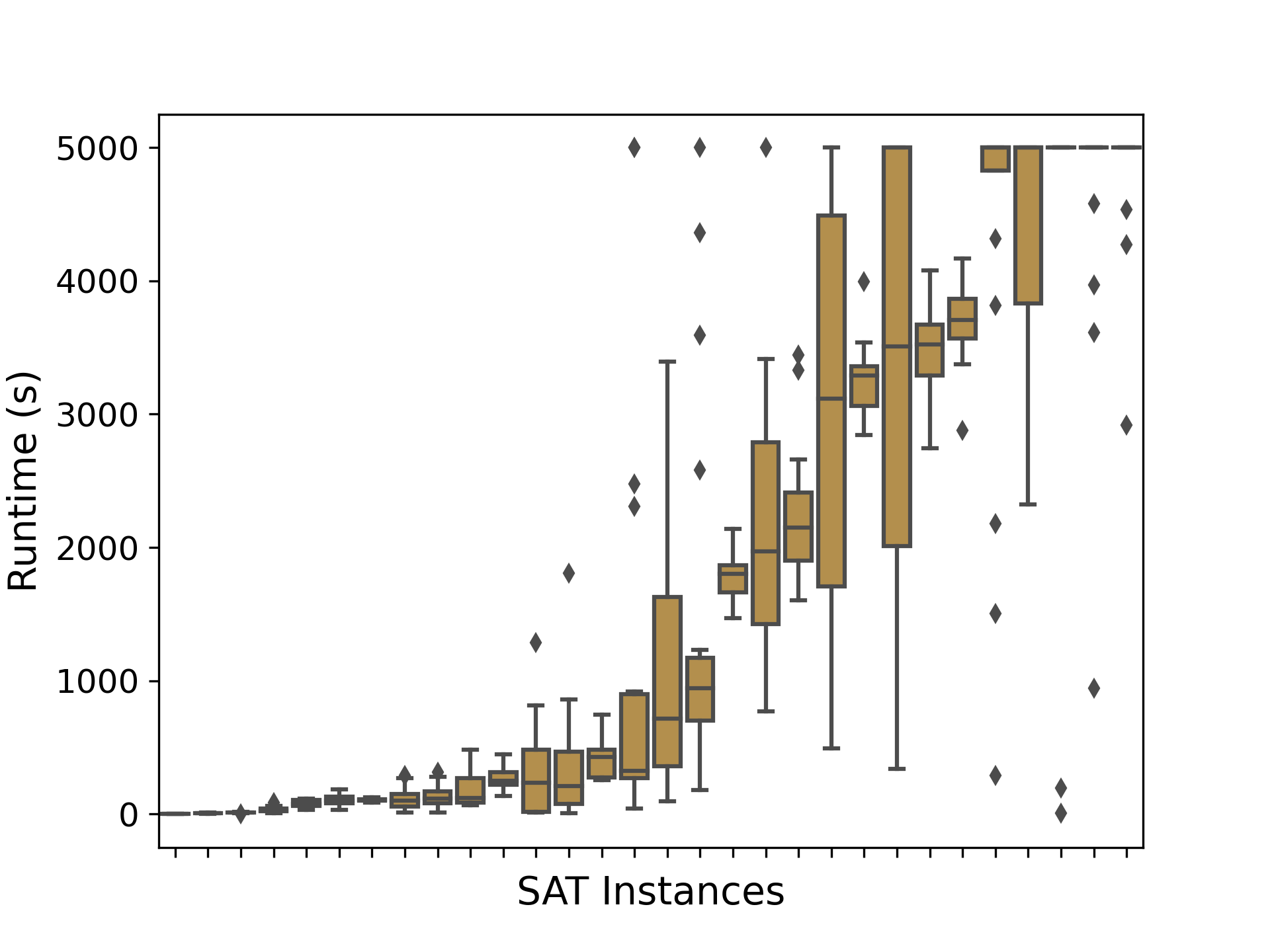}
    \end{subfigure}
\caption{Runtime variation introduced by permutation. Thirty SAT instances are randomly sampled from SAT Competition data, and the order of variables (left) and clauses (right) are shuffled twenty times. The instances are then solved by Kissat 3.0 with a 5,000s cutoff, and the collected runtimes are plotted in ascending order of mean runtime.}
\label{fig:permutation}
\end{figure*}

In principle, satisfiability of a SAT formula is not affected by permuting the variables or clauses, and literal-clause graphs are permutation-invariant as well. In practice, however, we found solver runtimes can be sensitive to the order in which clauses are provided as input. We conducted a study with the popular Kissat 3.0 \citep{SAT2022solvers} solver on the industrial LEC dataset described further in Section~\ref{sec:experiments}. As can be seen in Figure~\ref{fig:permutation}, shuffling clauses sometimes led to very large variations in runtime. In contrast, shuffling variables showed limited impact.

This is in line with previously reported remarks on other SAT solvers \citep{durairaj2005variable,clause-order}. A possible explanation could be the algorithmic design of modern solvers, for which the storage architecture of variables relies on a doubly linked list and the initial storage order follows the parsing order of the clauses. This results in variations in cache miss rates depending on the provided clause ordering. In contrast, variable ordering usually only impacts the variable labels used by the solvers. 

To address the sensitivity to clause ordering, we include positional encodings among the clause features. These encode the position of a clause within the CNF formula. We follow the classical encodings from \citet{vaswani2017attention} and endow the $k^\text{th}$ clause with a 10-dimensional embedding
\begin{align*}
    PE(k, 2i) &= \sin\bigg( \frac{k}{10000^{2i/10}}\bigg), \\
    PE(k, 2i\!+\!1) &= \cos\bigg( \frac{k}{10000^{2i/10}}\bigg),
\end{align*}
where $i=0,\dots,4$. This vector is concatenated with the rest of the clause node features.

\subsection{Model}

We use a graph neural network (GNN) model to predict which solver to use for a given instance. These models operate on graphs by repeatedly modifying node embeddings through graph convolution operations, and have emerged as a standard paradigm for dealing with graph-structured data, both in SAT solving \citep{guo2023machine} and more widely for combinatorial optimization in general \citep{cappart2023combinatorial}. However, we deviate from standard graph convolution frameworks by interpreting our literal-clause graph as a graph with three types of edges: (i) from clause to literal nodes; (ii) from literal to clauses nodes; and (iii) between positive and negative literal nodes. These graphs can then be interpreted as ``heterogeneous graphs'', and we can apply heterogenous GNN methodologies~\citep{zhang2019heterogeneous}.

In this framework, the graph convolution steps take the following edge-type-dependent form. Let $x_j\in\mathbb{R}^d$ stand for the feature vector at node $j$. For every edge $(j,i)\in\mathcal{E}_k$ of type $k\in\{1,\dots,K\}$ from node $j$ to node $i$, we compute a message
$m_{i, j, k} = \phi_k(x_i, x_j)$ where $\phi_k$ is a learnable ``message function''. We then update the feature vector at node $i$ by the formula
\begin{align*}
    \overline{m}_{i,k} =&\; \rho_k \big(\big\{m_{i, j, k} \,\big\vert\, (j, i) \in \mathcal{E}_k \big\}\big), \\
    x_{i} \leftarrow&\; \delta\Big(\Big\{\,\psi_k \big(  x_i,\,\overline{m}_{i,k}\big)\,\Big\vert\, k=1\dots,K\,\Big\}\Big)\,,
\end{align*}
where $\rho_k$ is the ``message aggregation'' function for edge type $k$, $\psi_k$ is the learnable ``update'' function, and $\delta$ is the ``edge-type aggregation'' function.

We apply this framework to our neural network as follows. For clarify, a diagram of the architecture is provided as Figure 1. The clause, positive and negative literal node embeddings are initialized with the 17-, 3-, and 3-dimensional feature vectors described in Table \ref{tab:node-features},
\begin{align*}
x_{\text{clause},i}&\leftarrow \text{features}_{\text{clause},i}\;\;\;\forall\text{ clause node }i\\
x_{\text{poslit},i}&\leftarrow \text{features}_{\text{poslit},i}\quad\forall\text{ positive literal node }i\\
x_{\text{neglit},i}&\leftarrow \text{features}_{\text{neglit},i}\quad\forall\text{ negative literal node }i
\end{align*}

These are then transformed by two rounds of convolutions. We set $\phi_k$ and $\psi_k$ as in the classical graph convolutions of \citet{kipf2017semisupervised}, with a relu nonlinearity and mean aggregation functions for $\rho_k$ and $\delta$. That is, we compute

\begin{align*}
x_{\text{clause},i}\leftarrow 
\text{relu}\Big(b &+\sum_{j\in N_\text{poslit}(i)}\frac{W_\text{poslit}x_{\text{poslit},j}}{\sqrt{\text{deg}(i)\text{deg}(j)}} \\
&+\sum_{j\in N_\text{poslit}(i)}\frac{W_\text{neglit}x_{\text{neglit},j}}{\sqrt{\text{deg}(i)\text{deg}(j)}}\Big)
\end{align*}

\subsection{Training}

We train in a supervised fashion on a dataset of training SAT instances, for which runtimes have been collected ahead of time on each solver of interest. Since our model produces a distribution over the $K$ possible solvers, we could treat the problem as simple classification with a cross-entropy loss. However, this would not be well-aligned with our objective of minimizing solving runtime, since it would equally penalize incorrect predictions, irrespective of the amount of additional runtime induced by the selection of a suboptimal solver. Instead, we propose the regret-like loss
\begin{equation}
    \mathcal{L} = \frac{1}{N}\sum_{i=1}^N \bigg( \sum_{k=1}^K p_i^k t_i^k - t_{i}^{*} \bigg)^2\,,
\end{equation}
where $p_i^k$ and $t_i^k$ are the model probability and runtime for instance $i=1,\dots,N$ and solver $k=1,\dots,K$, respectively, and $t_{i}^{*}=\min_k t_i^k$ is the best time achieved by any solver on the instance. This has the advantage of more directly optimizing final runtime, taking into account that not all mistakes are equally impactful on solving time. We minimize this loss using the Adam \citep{kingma2014adam} algorithm with early stopping.

\subsection{Inference}

At test-time, to predict which SAT solver to use, we convert the SAT instance into our graph representation, compute its node features, and feed the graph to our trained GNN model, which outputs a probability distribution over the solvers of interest. The solver with highest probability is chosen for solving the instance, with the runtime being reported.
\section{Experiments}\label{sec:experiments}
We now compare the performance of our proposed approach against competitors in the literature. We implement our model using the \texttt{pytorch} \citep{paszke2019pytorch} and \texttt{dgl} \citep{wang2019deep} libraries. We train the model on a single Nvidia Tesla V100 GPU with a learning rate of $1e{-}3$ for up to 100 epochs, and use the same GPU at test-time.

\subsection{Base solvers}\label{sec:base-solvers}

We train and evaluate on a portfolio of seven top-performing solvers from recent SAT Competitions \citep{SAT2020proceeding, SAT2022solvers}: (a) \texttt{Kissat-3.0}, (b) \texttt{bulky}, (c) \texttt{HyWalk}, (d) \texttt{MOSS}, (e) \texttt{mabgb}, (f) \texttt{ESA} and (g) \texttt{UCB}. Among them, \texttt{Kissat-3.0} and \texttt{bulky} are based on \texttt{Kissat}, which is the winner of the 2020 SAT Competition and is known for its efficient data structure design \citep{SAT2020proceeding}. The other five solvers are based on \texttt{UCB} and differ in their bandit-based scoring mechanism for branching. It should be noted that our method works for any choice of candidate solvers.

\begin{table}[t]
\centering
\caption{Dataset statistics. We report the average number of instances, variables and clauses, for different groups of instances.}
\begin{subtable}{\columnwidth}
{\resizebox{\columnwidth}{!}{
\begin{tabular}{@{}lrrr|lrrr@{}}
\toprule
Name       & \# instances & \# variables & \# clauses & Name       & \# instances & \# variables & \# clauses \\
\midrule
Circuit 1  & 788    & 10,404 & 39,502    & Circuit 16 & 22,050 & 1,002  & 3,969     \\
Circuit 2  & 49     & 6,199  & 26,108    & Circuit 17 & 10     & 1,140  & 3,914     \\
Circuit 3  & 36     & 3,612  & 13,079    & Circuit 18 & 798    & 1,058  & 3,829     \\
Circuit 4  & 2,014  & 1,783  & 7,074     & Circuit 19 & 38     & 1,067  & 3,770     \\
Circuit 5  & 21     & 2,421  & 6,830     & Circuit 20 & 14,840 & 952    & 3,663     \\
Circuit 6  & 3,823  & 1,559  & 6,281     & Circuit 21 & 2,024  & 929    & 3,593     \\
Circuit 7  & 2      & 401    & 6,172     & Circuit 22 & 23     & 1,040  & 3,256     \\
Circuit 8  & 25,591 & 1,563  & 6,171     & Circuit 23 & 401    & 932    & 3,085     \\
Circuit 9  & 615    & 1,639  & 6,034     & Circuit 24 & 105    & 909    & 2,297     \\
Circuit 10 & 582    & 1,709  & 5,940     & Circuit 25 & 1,420  & 851    & 2,109     \\
Circuit 11 & 736    & 1,412  & 5,308     & Circuit 26 & 69     & 627    & 1,658     \\
Circuit 12 & 99     & 1,341  & 5,246     & Circuit 27 & 16     & 497    & 1,418     \\
Circuit 13 & 56     & 1,540  & 5,230     & Circuit 28 & 298    & 454    & 1,392     \\
Circuit 14 & 978    & 1,224  & 4,757     & Circuit 29 & 564    & 411    & 1,389     \\
Circuit 15 & 662    & 1,051  & 4,109     & Circuit 30 & 21     & 491    & 1,165     \\
\bottomrule
\end{tabular}
}}\vspace{5pt}
\caption{LEC data. The instances are regrouped by the circuit optimization sequence from which they were generated.}
\label{tab:lec_data}
\vspace{10pt}
\end{subtable}
\begin{subtable}{\columnwidth}
\begin{tabularx}{\columnwidth}{@{ }lrrrr@{}}
\toprule
Runtime Range (s)\hspace{27pt} & \# instances    & \# variables & \# clauses \\ \midrule
(0, 1{]}      & 747             & 2,505         & 48,163      \\
(1, 10{]}     & 416             & 4,212         & 74,203      \\
(10, 100{]}   & 427             & 4,415         & 98,151      \\
(100, 1500{]} & 498             & 4,250         & 135,969     \\ \bottomrule
\end{tabularx}\vspace{5pt}
\caption{SAT Competition data. The instances are regrouped by their best runtime among the base solvers of Subsection \ref{sec:base-solvers}.}
\label{tab:sc_data}
\end{subtable}
\end{table}

\subsection{Datasets}
We train and evaluate on two datasets. 

\paragraph{Logic Equivalence Checking (LEC)} This is a proprietary dataset generated from logic equivalence checking steps in electronic circuit design.  Circuits undergo a large number of optimization steps during logic synthesis, and at each one of the steps, it is necessary to verify that the circuits before and after optimization are functionally equivalent. This is done by verifying that the two circuits produce the same outputs for all possible inputs, which is equivalent to solving a SAT problem \citep{huang2012formal, goldberg2001using}. We collected logic equivalence checks from the optimization of 30 industrial circuits, yielding a total of 78,727 SAT instances. A summary of dataset statistics is provided as Table \ref{tab:lec_data}.

\paragraph{SAT Competition (SC)} This is a subset of the Anniversary Track Benchmark of the 2022 SAT Competition~\citep{satcompetition2022}, which itself was created by collecting all instances from the Main, Crafted and Application tracks of the previous SAT competitions up to that year. We ran each instance of the Anniversary benchmark through each of the seven solvers in the portfolio, and excluded those that could not be solved within 1,500 seconds by any solver, as well as those with more than 20,000 variables, yielding 2,088 SAT instances. A summary of dataset statistics is provided as Table \ref{tab:sc_data}.

\subsection{Baselines}

We compare our approach with the following baselines.

\paragraph{Best Base Solver} The individual solver among the portfolio of seven that had the best performance on the training data, measured in average runtime over all instances. In practice, this was the \texttt{bulky} solver for both datasets.

\paragraph{SATzilla07 \citep{xu2008SATzilla}} We adapt this landmark SAT solving machine learning model, based on a linear ridge regression model trained to predict runtimes based on global handcrafted features that summarize SAT instance characteristics. Since our work focuses on SAT solver selection, we omit the presolving process in the original SATzilla pipeline. We also remove features in the original model that require probing. This leaves 33 global features (\#1-33 in the original article). The model is trained from the \texttt{Ridge} class in the \texttt{scikit-learn} \citep{pedregosa2011scikit} library with default settings. We convert the approach into a SAT solver selection model by selecting the solver with shortest predicted runtime.

\paragraph{SATzilla12 \citep{xu2012satzilla}} We also adapt the updated SATzilla model from 2012, which was based on random forest classification. Again, we remove the presolving process, and only use the features that do not require probing (features \#1-55 in the original article). We train a random forest model between each pair of solvers, weighting each training instance by the absolute difference in runtime between the two solvers, for $7(7-1)/2=21$ models in total. Each model is trained from the \texttt{RandomForestClassifier} class in \texttt{scikit-learn} with 99 trees and 
$\lceil log_2(55)\rceil+1 = 7$ sampled features in each tree. At test-time, each model is used to vote which solver it prefers in its pair for solving an instance, and the final solver choice is made from the solver that has received the most votes.

\paragraph{ArgoSmArT \citep{nikolic2013argosmart}} This is an approach based on a $k$-nearest neighbors model trained for classification. We used the same 29 features as in the original paper, which form a subset of the 33 features used in our adaptation of SATzilla07. We use the \texttt{KNeighborsClassifier} class in \texttt{scikit-learn} with $k=9$ neighbors.

\paragraph{CNN \citep{loreggia2016deep}} We reimplement the approach of \citet{loreggia2016deep}, where the CNF formula is interpreted as text, converted to its ASCII values and then to a grayscale image, before being resized to 128x128 pixels and fed to a Convolutional Neural Network (CNN). We use the same architecture as in the paper, which we implement in the \texttt{pytorch} \citep{paszke2019pytorch} library, and train it over a cross-entropy loss with the Adam \cite{kingma2014adam} algorithm, a learning rate of 1e-3 and early stopping.

\subsection{Metrics}\label{sec:metrics}

We report the following metrics for performance evaluation. Results are averaged over five train-test folds over the data, and the average and standard deviation over those five folds are reported.

\paragraph{Average Runtime (Avg Runtime)} For each instance, the method selects a solver, and this solver is used to solve the instance, reporting a runtime. These runtimes are then averaged over all instances. In other words, this is the average runtime that would be observed if this method were used for all instances, as a ``portfolio" solver. Lower is better.

\paragraph{Percentage of Solved Instances (Solved)} The percentage of instances that can be solved by the selected solver within a cutoff time of 500s. Higher is better.

\paragraph{Classification Accuracy (ACC)} The accuracy of selecting the optimal solver for a given instance. This measures how efficient a method is in selecting the optimal solver. Higher is better.

\begin{table}[t]
\caption{Main results on the LEC and SC benchmarks. We report the average and standard deviation over 5 train-test folds. For every best result, we bold the number and conduct a Wilcoxon signed-rank test to test whether the distribution of the differences in results between this method and the next best method for every instance and fold is equally distributed around zero. An asterisk (${}^*$) next to the number denotes a p-value lower than 0.05. ${}^\dag$The CNN on the LEC dataset predicted \texttt{bulky} for every instance, giving identical results to "Best base solver".}
\begin{tabularx}{\columnwidth}{@{}l>{\hspace{-2pt}}rrr@{}}
\toprule
 & \hspace{-40pt}Avg Runtime (s) $\downarrow$    & Solved (\%) $\uparrow$ & ACC $\uparrow$        \\[2pt] \toprule
\multicolumn{4}{c}{LEC} \\[-1pt] \midrule
Best base solver    & 382.783±1.877             & 74.7±0.2 & 0.424±0.001\\
SATzilla07          & 346.492±1.516             & 76.8±0.2 & 0.467±0.001\\
SATzilla12          & 344.290±1.115              & 77.2±0.1 & ${}^*$\textbf{0.487}±0.004\\
ArgoSmArT           & 353.241±1.093             & 76.3±0.2 & 0.457±0.002\\
CNN$^\dag$          & 382.783±1.877             & 74.7±0.2 & 0.424±0.001\\ %
GraSS (ours)        & ${}^*$\textbf{341.549}±1.440    & ${}^*$\textbf{77.7}±0.2 & 0.480±0.006  \\ \toprule %
\multicolumn{4}{c}{SC} \\[-1pt] \midrule
Best base solver    & 250.902±21.669            & 82.9±1.7 & 0.210±0.013\\ %
SATzilla07          & 227.643±11.955            & 83.5±2.2 & 0.330±0.018\\ %
SATzilla12          & 222.146±18.311            & 84.0±2.0 & 0.366±0.009\\ %
ArgoSmArT           & 227.121±14.312            & 83.9±1.7 & ${}^*$\textbf{0.449}±0.023\\ %
CNN                 & 253.832±11.985            & 82.4±0.8 & 0.296±0.016  \\ %
GraSS (ours)        & ${}^*$\textbf{220.251}±16.360   & \textbf{84.6}±1.8 & 0.259±0.026\\ \bottomrule
\end{tabularx}
\label{tab:main_table}
\end{table}

\begin{table}[t]
\caption{Detailed comparison of the performance of GraSS and the next best method, SATzilla12, over best-runtime quantiles. Results are averaged over 5 folds, and measured in seconds.}
\begin{tabularx}{\columnwidth}{@{}lrrrr}
\toprule
\multirow{2}{2.2cm}{Best-Runtime Quantile} & \multicolumn{2}{c}{LEC} & \multicolumn{2}{c}{SC} \\
\cmidrule(lr){2-3}\cmidrule(lr){4-5}
               & SATzilla12 & GraSS   & SATzilla12 & GraSS   \\[1pt] \toprule
{[}0, 0.25{]}  & \textbf{103.387} & 104.575                 & \textbf{0.440} & 0.673 \\
(0.25, 0.50{]} & 231.500          & \textbf{229.448}          & \textbf{14.541} & 18.691 \\
(0.50, 0.75{]} & 390.999          & \textbf{386.567}          & 143.847 & \textbf{142.970} \\
(0.75, 1{]}    & 653.293          & \textbf{647.625}          & 736.697 & \textbf{721.190} \\ \bottomrule
\end{tabularx}
\label{tab:quantile_res}
\end{table}

\begin{figure}
    \centering
    \begin{subfigure}{\columnwidth}
    \includegraphics[width=\columnwidth]{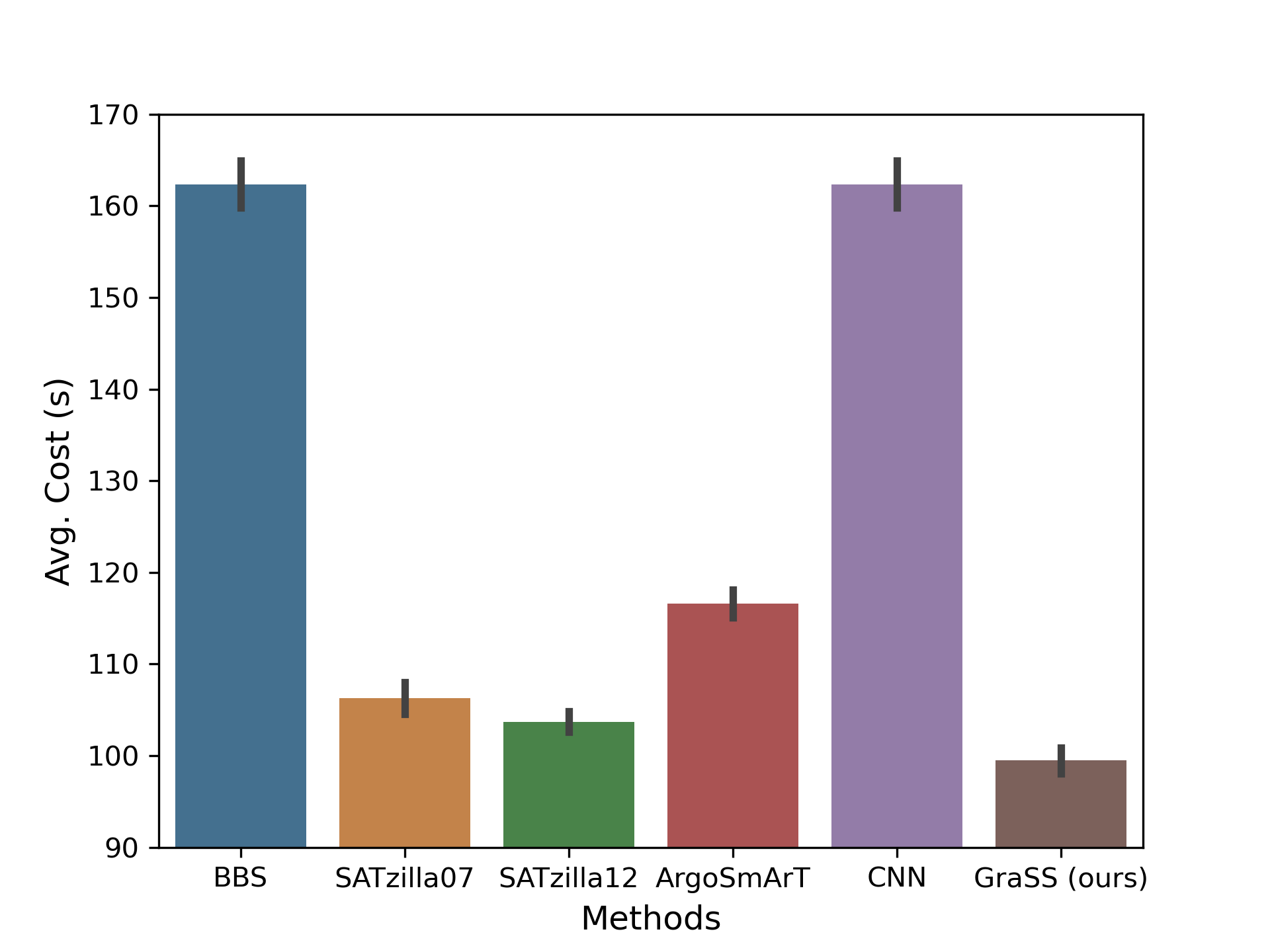}
    \caption{LEC benchmark.}
    \end{subfigure}
    \begin{subfigure}{\columnwidth}
    \includegraphics[width=\columnwidth]{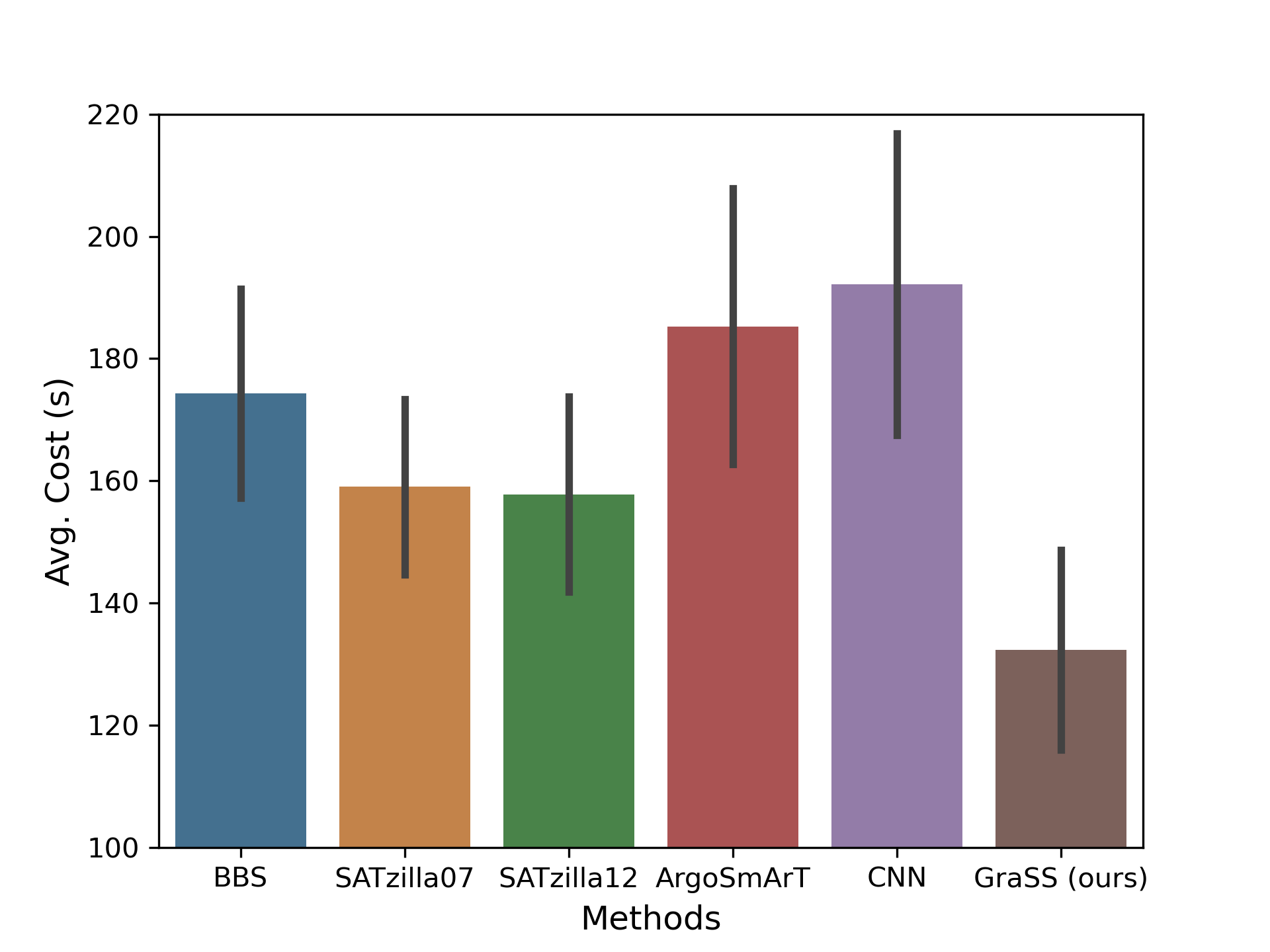}
    \caption{SC benchmark.}
    \end{subfigure}
    \caption{Cost of wrong prediction: the runtime difference between predicted solver and the optimal solver, when the selector has made a mistake. Average across five folds are shown, with standard deviation as error bar. Lower is better.}
    \label{fig:cost_wrong}
\end{figure}

\subsection{Main Results}

We report in Table~\ref{tab:main_table} the main results of our experiments on the Logic Equivalence Checking (LEC) and SAT Competition (SAT) datasets, respectively. As can be seen, our proposed GraSS method outperform competing approaches in average runtime, as well as in percentage of problems solved within our 500s cutoff time. Interestingly, this is true despite the method not being as accurate in selecting the optimal solver for every instance, as measured by the accuracy metric. This suggests that an important component of its success lies in its improved robustness to error: when the method makes mistakes, they impact runtime less than competing methods.

To understand this phenomenon further, we looked at the runtime difference between the predicted solver and the optimal solver, whenever a mistake is made. As can be seen in Figure \ref{fig:cost_wrong}, the average cost of wrong prediction is substantially lower for our method than for competitors, especially for the SC dataset.

We further analyze the performance difference between our approach and the next best method in average runtime, SATzilla12. Table \ref{tab:quantile_res} regroups the instances by rough measure of difficulty, namely by grouping them by which quartile (0-25\%, 25-50\%, 50-75\% or 75-100\%) their best runtime achieved on any solver falls into, and compares the performance of SATzilla12 and GraSS on each subgroup. As can be seen, SATzilla12 performs better on easy instances, while GraSS performs better on hard instances.

Finally, as described in Subsection \ref{sec:metrics}, our timing results only report the time taken by the solvers in optimizing the instances. In particular, this means we exclude from the numbers the time taken to compute the features necessary to take the decision, which is dependent on the storage format used to save the instances. For our experiments, we chose to save them on a hard drive in the standard \texttt{.cnf} text file format \citep{cnf}, and we report for completeness the time taken to compute the features for each method in Table \ref{tab:preprocess_time}. Other storage formats would lead to different timings.

\begin{table}[t]
\caption{Comparison of the time taken to compute the features required for each method from \texttt{.cnf} files. We report the mean and standard deviation in seconds over the instances for each dataset.}
\label{tab:preprocess_time}
\begin{tabularx}{\columnwidth}{@{}lrp{0.5cm}r@{}}
\toprule
\hspace{0.61\columnwidth} & \multicolumn{1}{c}{LEC\hspace{5pt}}  & &\multicolumn{1}{c}{\;\;SC}      \\[1pt] \toprule
Best base solver & \multicolumn{1}{c}{--} & &\multicolumn{1}{c}{\;\;--} \\
SATzilla07  & 6.767 && 35.932  \\
SATzilla12  & 8.643 && 194.155 \\
ArgoSmArT   & 6.472 && 41.266 \\
CNN         & 1.353 && \textbf{2.401}   \\
GraSS (ours) & \textbf{1.232}&& 33.862  \\ \bottomrule
\end{tabularx}
\end{table}

\subsection{Ablation Study}

We extend our analysis by considering the impact of various methodological choices on performance over the LEC benchmark.

We first evaluate the impact of the graph neural network architecture, by comparing our approach with a variant that uses the same convolution weights for every edge, effectively treating it as a homogeneous graph (\emph{Homogeneous}). We also compare with a NeuroSAT-style architecture (\emph{NeuroSAT variant}) inspired by \citet{selsam2019neurosat}, which was originally designed for satisfiability prediction (sat/unsat). Their model also uses a literal-clause graph to encode instances, although with learned initial node embeddings, and uses a very deep LSTM-GNN hybrid architecture with 26 layers and custom graph convolution operations. We implement the same, but replace the final layer which computes a scalar ``vote'' for every literal, and takes the average vote before a sigmoid activation, by an averaging of the literal embeddings, followed by a linear layer and a softmax activation. We also use 4 layers instead of 26 for tractability on our dataset, whose instances are substantially larger than those in the original paper. As can be seen in Table \ref{tab:graph_struct}, our approach improves over these alternatives in every metric.

We next evaluate our choice of node features. We compare it against random normal values (\emph{Random}), as in \citet{selsam2019neurosat}); a one-hot vector indicating whether the node represents a clause, positive or negative literal (\emph{Node-type}), as used in \citet{yolcu2019nips, li2023hardsatgen} and \citet{you2019g2sat}; and Laplacian Positional Encodings (\emph{Laplacian PE}), as introduced in \citet{dwivedi2023benchmarking}. We also compare against a variant of our approach consisting of the same hand-designed features, but without the clause positional embeddings (\emph{Custom}). As can be seen in Table \ref{tab:ablation-features}, our choices outperform these alternative approaches in every metric.
\section{Limitations}

Although our experiments strongly establish the superiority of our approach in the presented scenario, several limitations can be noted.
Deep learning methods are well-known to be data hungry, and perform best in regimes where training sets are large. It is plausible that in scenarios where a limited number of timed instances are available, performance would not be competitive against simpler models. In addition, in many scenarios it might be desirable to learn online, updating models as examples stream in: our method cannot be readily adapted to this situation, as training requires runtime labels on every solver for each instance, and adapting graph neural networks to online learning is challenging \citep{wang2020streaming}.

\begin{table}[t]
\caption{Exploration of alternative architectures on the LEC benchmark. We report the average and standard deviation over 5 train-test folds.}
\begin{tabularx}{\columnwidth}{@{}lrrr@{}}
\toprule
 & \hspace{-12pt}Avg Runtime (s) $\downarrow$   & Solved (\%) $\uparrow$ & ACC $\uparrow$ \\[2pt] \toprule
Homogeneous         & 343.339±0.987             & 77.4±0.2 & 0.464±0.009\\
NeuroSAT variant    & 383.132±5.284             & 74.3±1.0 & 0.423±0.008\\
GraSS (ours)        & \textbf{341.549}±1.440    & \textbf{77.7}±0.2  & \textbf{0.480}±0.006\\ \bottomrule
\end{tabularx}
\label{tab:graph_struct}
\end{table}

\begin{table}[t]
\caption{Exploration of alternative node features on the LEC benchmark. We report the average and standard deviation over 5 train-test folds.}
\begin{tabularx}{\columnwidth}{@{}lrrr@{}}
\toprule
& \hspace{-20pt}Avg Runtime (s) $\downarrow$     & Solved (\%) $\uparrow$ & ACC $\uparrow$        \\[2pt] \toprule
Random              & 352.258±2.114 & 76.2±0.5 & 0.444±0.008\\
Node-type           & 344.088±1.603 & 77.1±0.3 & 0.474±0.002\\
Laplacian PE        & 347.632±1.274 & 76.9±0.3 & 0.454±0.003\\ 
Custom              & 343.143±1.621 & 77.3±0.3 & 0.476±0.003\\
Custom + PE (ours)  \hspace{-7pt} & \textbf{341.549}±1.440 & \textbf{77.7}±0.2 & \textbf{0.480}±0.006\\
\bottomrule
\end{tabularx}
\label{tab:ablation-features}
\end{table}
\section{Conclusion}\label{sec:conclusion}

This work proposed a novel supervised approach to SAT solver selection, based on representing instances as literal-clause graphs and training a graph neural network to select, from this representation, a SAT solver among a fixed portfolio so as to minimize solving runtime. The graph representations are endowed with node features that encode domain knowledge, and in the case of clause nodes, also position within the SAT formula. The resulting scheme is shown to outperform competing approaches on two benchmarks, one from an industrial circuit design application and one from the annual SAT solver competitions.

\bibliography{main.bib}
\bibliographystyle{ACM-Reference-Format}

\newpage
\appendix
\section{Node features}\label{sec:node-features}

We provide in Table \ref{tab:node-features} a summary of the features in our graph representations, for each node type.

\begin{table}[t]
\caption{Node features used in our graphical representation of a SAT instance. We report the feature dimension, feature name and a description.}
\begin{tabular}{rlp{113pt}}
\toprule
\multicolumn{3}{c}{\bf Positive literal node features} \\
\midrule
1 & \texttt{pos\_lit\_degree} & Number of clauses the positive literal appears in, divided by the total number of clauses. \\
1 & \texttt{pos\_lit\_num\_horn} & Number of Horn clauses the positive literal appears in, divided by the total number of clauses.\\
1 & \texttt{lit\_pos\_neg\_ratio} & Number of clauses the positive literal appears in, divided by the number of clauses its negation appears in plus 1. \\
\midrule
\multicolumn{3}{c}{\bf Negative literal node features} \\
\midrule
1 & \texttt{neg\_lit\_degree} & Number of clauses the negative literal appears in, divided by the total number of clauses. \\
1 & \texttt{neg\_lit\_num\_horn} & Number of Horn clauses the negative literal appears in, divided by the total number of clauses.\\
1 & \texttt{lit\_pos\_neg\_ratio} & Number of clauses the positive literal appears in, divided by the number of clauses its negation appears in plus 1. \\
\midrule
\multicolumn{3}{c}{\bf Clause node features} \\
\midrule
1 & \texttt{clause\_is\_horn} & Is the clause Horn?  \\
1 & \texttt{clause\_degree} & Number of literals in the clause, divided by the total number of variables in the instance. \\
1 & \texttt{clause\_is\_binary} & Is the clause composed of two literals?  \\
1 & \texttt{clause\_is\_ternary} & Is the clause composed of three literals? \\
1 & \texttt{clause\_pos\_num} & Number of positive literals divided by the total number of literals in the clause. \\
1 & \texttt{clause\_neg\_num} & Number of negative literals divided by the total number of literals in the clause. \\
1 & \texttt{clause\_pos\_neg\_ratio} & Number of postivive literals, divided by the number of negative literals in the clause plus 1. \\
10 & \texttt{clause\_pe} & Positional encoding (see Subsection \ref{sec:clause_pe} in the main paper.) \\
\bottomrule
\end{tabular}
\label{tab:node-features}
\end{table}

\end{document}